\documentclass{article}
\pdfoutput=1
\usepackage[utf8]{inputenc} % allow utf-8 input
\usepackage[T1]{fontenc}    % use 8-bit T1 fonts
\usepackage[numbers]{natbib}
\setcitestyle{numbers,square}
\usepackage{appendix}
\usepackage{tabularx}

\usepackage[table]{xcolor} 
\usepackage{multirow}

\definecolor{tableGray}{gray}{0.92} 
\definecolor{tableBlue}{HTML}{E8F6FF}  
\usepackage[preprint]{neurips_2024}
\usepackage{hyperref}
\usepackage{graphicx}
\usepackage{amsmath}
\usepackage{makecell}
\usepackage{easyReview}
\usepackage{threeparttable}   
\usepackage{amsfonts}       % blackboard math symbols

\usepackage{xcolor}         % colors

\usepackage{multirow}
\usepackage{enumitem}

\usepackage{amsthm}
\usepackage{mathtools}

\definecolor{mutedRed}{HTML}{F95454}
\definecolor{mutedBlue}{HTML}{0D92F4}

\definecolor{brightorange}{RGB}{255, 165, 0}
\definecolor{newpurple}{RGB}{64, 3, 156}

\usepackage[utf8]{inputenc} % allow utf-8 input
\usepackage[T1]{fontenc}    % use 8-bit T1 fonts
\usepackage{booktabs}       % professional-quality tables
\usepackage{amsfonts}       % blackboard math symbols
\usepackage{nicefrac}       % compact symbols for 1/2, etc.
\usepackage{microtype}      % microtypography
\title{GSTM-HMU: Generative Spatio-Temporal Modeling for Human Mobility Understanding}

\author{%
  Wenying Luo\textsuperscript{\rm 1},
  ~Zhiyuan Lin\textsuperscript{\rm 1,2},
  ~Wenhao Xu\textsuperscript{\rm 1,2},
  ~Minghao Liu\textsuperscript{\rm 2},
  ~Zhi Li\textsuperscript{\rm 1}\\
  \textsuperscript{\rm 1}City University of Hong Kong \\
  \textsuperscript{\rm 2}Nankai University \\
}

\begin{document}

\maketitle
\begin{abstract}
Human mobility traces, often recorded as sequences of check-ins, provide a unique window into both short-term visiting patterns and persistent lifestyle regularities. In this work we introduce GSTM-HMU, a generative spatio-temporal framework designed to advance mobility analysis by explicitly modeling the semantic and temporal complexity of human movement. The framework consists of four key innovations. First, a Spatio-Temporal Concept Encoder (STCE) integrates geographic location, POI category semantics, and periodic temporal rhythms into unified vector representations. Second, a Cognitive Trajectory Memory (CTM) adaptively filters historical visits, emphasizing recent and behaviorally salient events in order to capture user intent more effectively. Third, a Lifestyle Concept Bank (LCB) contributes structured human preference cues, such as activity types and lifestyle patterns, to enhance interpretability and personalization. Finally, task-oriented generative heads transform the learned representations into predictions for multiple downstream tasks. We conduct extensive experiments on four widely used real-world datasets, including Gowalla, WeePlace, Brightkite, and FourSquare, and evaluate performance on three benchmark tasks: next-location prediction, trajectory-user identification, and time estimation. The results demonstrate consistent and substantial improvements over strong baselines, confirming the effectiveness of GSTM-HMU in extracting semantic regularities from complex mobility data. Beyond raw performance gains, our findings also suggest that generative modeling provides a promising foundation for building more robust, interpretable, and generalizable systems for human mobility intelligence.
\end{abstract}
% \vspace{-4mm}
\section{Introduction}
% \vspace{-3mm}
\label{Introduction}
\label{section:intro}
Location-based services (LBS) such as Gowalla, Weeplace, and Foursquare have generated an unprecedented volume of human mobility data, commonly represented as sequences of check-ins. Each check-in corresponds to a visit to a point of interest (POI), such as restaurants, hospitals, or recreational venues, and is accompanied by temporal and geographical metadata. These check-in traces are not only digital footprints of human activities but also encode meaningful semantics that reflect user intentions, behavioral patterns, and lifestyle preferences. Understanding such semantics is vital for urban computing, personalized recommendations, and sustainable city planning~\cite{zhang2014human, yuan2018deepmob, feng2020deepmove}.

The central challenge in mining check-in sequences lies in extracting multi-level semantics beyond surface-level trajectories. Previous studies mainly emphasized task-specific predictions, such as next location forecasting~\cite{kong2018hst, feng2020deepmove, liao2021deep}, time-of-arrival prediction~\cite{zuo2021tale, xu2022ifl}, and trajectory-user linking~\cite{zhao2018tuler, guo2020tulvae, xu2021deepsim}. While effective, these approaches often underexplore the latent behavioral drivers, leading to limited generalization. More recently, large language models (LLMs) have shown remarkable ability in semantic abstraction and contextual reasoning across diverse modalities~\cite{brown2020gpt3, openai2023gpt4, liu2023llm4ts, 10.1145/3664647.3681115}. This inspires a new perspective: mobility data can be reprogrammed into semantic sequences interpretable by generative models.

However, directly applying LLMs to mobility data faces critical obstacles. Unlike natural text, check-in sequences intertwine spatial, temporal, and categorical information, which require specialized encoding to preserve structure. Moreover, mobility signals often comprise two intertwined components: (i) short-term dynamics, reflecting immediate intentions (e.g., commuting after work), and (ii) long-term regularities, reflecting lifestyle and preferences (e.g., frequent visits to gyms or cafes). Capturing both simultaneously remains challenging.

To overcome these limitations, we propose a novel framework named GSTM-HMU, designed as a generative spatio-temporal learner for check-in sequence analysis. Unlike previous works, GSTM-HMU integrates multiple innovations:
\begin{itemize}[leftmargin=5mm]
\item We introduce a Spatio-Temporal Concept Encoder (STCE) that jointly embeds geographical coordinates, POI categories, and periodic time rhythms into unified semantic vectors, enabling the model to recognize both spatial context and temporal regularities.
\item A Cognitive Trajectory Memory (CTM) is designed to adaptively filter historical records using dual gating—prioritizing recency while highlighting atypical visits—thereby enhancing the extraction of user intentions.
\item A Lifestyle Concept Bank (LCB) provides domain-specific priors (e.g., occupation, activity, lifestyle) through semantic anchors. This mechanism acts as preference-aware prompts, aligning individual behaviors with broader lifestyle semantics~\cite{liu2023promptts, zhou2022conditional}.
\item GSTM-HMU employs generative task-oriented heads, predicting locations autoregressively, estimating time intervals with probabilistic decoding, and linking trajectories to users via pooled hidden states. This design facilitates flexible adaptation across tasks.
\item We validate our framework on four benchmark datasets and three downstream tasks. GSTM-HMU consistently outperforms recent baselines and shows robustness under few-shot training, suggesting its strong potential for real-world deployment.
\end{itemize}

In summary, GSTM-HMU reframes check-in sequence modeling as a generative spatio-temporal understanding problem, bridging LBS data mining and the recent advances in foundation models.

\section{Related Works}
% \vspace{-3mm}
\label{Related_Work}
Understanding human mobility requires bridging advances in trajectory analysis, generative sequence modeling, and cross-domain knowledge transfer. Below we highlight three complementary research lines most relevant to our work. A more comprehensive survey is deferred to Appendix.

\textbf{Trajectory Understanding and Spatio-Temporal Models.}  
Early studies focused on statistical heuristics for mobility forecasting~\cite{zheng2011urban, gao2012exploring}, while later works introduced neural sequence models to learn temporal dynamics. Recurrent architectures such as DeepMove~\cite{feng2020deepmove} and hierarchical attention designs~\cite{kong2018hst} improved next-location prediction, while contrastive methods~\cite{jiang2022will} explored representation robustness. Graph-enhanced frameworks, e.g., GNNTUL~\cite{GNNTUL}, extend to user identification by modeling trajectory–user dependencies. Despite these advances, existing solutions often remain task-specific, limiting cross-task generalization.

\textbf{Generative Foundations for Sequential Data.}  
The success of generative pre-trained transformers has motivated mobility researchers to reframe prediction tasks as sequence generation. Approaches such as GETNext~\cite{GETNext} and DualSin~\cite{chen2020dualsin} highlight autoregressive forecasting of POI sequences, while temporal point process models (e.g., SAHP~\cite{SAHP}, NSTPP~\cite{NSTPP}) emphasize fine-grained time estimation. Beyond mobility, One-Fits-All~\cite{One-fits-all} and AutoTimes~\cite{autotimes} show that a single frozen transformer can adapt to diverse sequential domains, hinting at the feasibility of mobility-specific generative backbones.

\textbf{Human-Centric Knowledge Integration.}  
Recent works emphasize injecting external or human-centered semantics into modeling. For instance, MoleculeSTM~\cite{MoleculeSTM} demonstrates how textual priors can guide molecular graphs, while LLM4TS~\cite{llm4ts} integrates large language models with time-series signals via multi-scale alignment. Similar ideas extend to visual and graph domains~\cite{lm4visual, TAPE}, underscoring that external prompts or anchors can bridge gaps between raw sequences and abstract semantics. In mobility, lifestyle-aware embeddings~\cite{zhang2024surveygenerativetechniquesspatialtemporal} have begun to capture long-term user preferences, but a unified framework that combines spatio-temporal encoding with semantic priors remains underexplored.

In contrast to these directions, our work seeks to unify trajectory modeling, generative reasoning, and human-centric priors into a single framework. We depart from isolated task formulations by rethinking check-in sequences as semantic narratives and designing a model that can simultaneously capture intentions, temporal rhythms, and lifestyle regularities.

\section{Preliminaries}
% \vspace{-2mm}
\paragraph{Data Universe and Notation.}
Let $\mathcal{U}$ denote the user set, $\mathcal{L}$ the set of points of interest (POIs), and $\mathcal{C}$ a taxonomy of semantically curated categories (e.g., \texttt{Food/Beverage} $\rightarrow$ \texttt{Cafe}).
Each check-in event is a marked tuple
$e=(\ell, t, \mathbf{g}, c, \mathbf{z})$ with location $\ell\in\mathcal{L}$, timestamp $t\in\mathbb{R}_{\ge 0}$, geodetic coordinate $\mathbf{g}\in\mathbb{S}^{2}$ (WGS84), category $c\in\mathcal{C}$, and optional context $\mathbf{z}$ (price tier, rating, device hints, etc.).
A user sequence (trajectory) is $C_u=\langle e_1,\ldots,e_n\rangle$ ordered by time.
For compactness we write $e_i=(\ell_i, t_i, \mathbf{g}_i, c_i, \mathbf{z}_i)$ and the history filtration $\mathcal{F}_{t} = \sigma(\{e_i: t_i \le t\})$.

\paragraph{Marked Temporal Point Process on the Sphere.}
We regard each $C_u$ as a \emph{marked temporal point process} on $(\mathbb{S}^{2}\times\mathbb{R}_{\ge 0},\mathcal{B})$ with counting measure $N_u(A)$ for $A\subseteq\mathbb{S}^{2}\times\mathbb{R}_{\ge 0}$.
The conditional intensity of the next event is
\[
\lambda_u(t, \mathbf{x}, c \mid \mathcal{F}_t)
\;=\;
\lambda_0(t)\,
\kappa_{\text{sp}}(\mathbf{x}\mid \mathcal{F}_t)\,
\kappa_{\text{cat}}(c\mid \mathcal{F}_t)\,
\psi(\Delta t \mid \mathcal{F}_t),
\]
where $\lambda_0(t)$ is a baseline, $\kappa_{\text{sp}}$ a spatial kernel on $\mathbb{S}^2$ (with geodesic distance $d_g$ via the haversine formula), $\kappa_{\text{cat}}$ a categorical compatibility, and $\psi$ an inter-event modulator; cf.\ Hawkes-style constructions and neural variants~\cite{mei2017nhp,zhang2020sahp}.

\paragraph{Spatial Discretization and Multi-Graphs.}
Exact spherical geometry is expensive at web scale. We adopt a hierarchical hexagonal indexer $h:\mathbb{S}^2\!\to\!\mathcal{H}$ (e.g., H3) to coarsen $\mathbf{g}$ into cells $h(\mathbf{g})$ with level $r$ controlling resolution.
We build heterogeneous graphs:
(i) $G_{\!L}=(\mathcal{L}, E_L)$ with edges for co-visit/proximity $w_{ij}=\exp(-d_g(\mathbf{g}_i,\mathbf{g}_j)/\tau)$,
(ii) $G_{\!C}=(\mathcal{C}, E_C)$ using the taxonomy,
(iii) $G_{\!H}=(\mathcal{H}, E_H)$ for cell adjacency.
Meta-paths (\emph{POI}\,$\to$\emph{Cell}\,$\to$\emph{POI}) enable structure-aware pooling. For meso-scale shape, we optionally compute persistent homology on sliding windows of coordinates to obtain topological summaries $\mathbf{p}_i$ (e.g., $\mathrm{H}_0,\mathrm{H}_1$ barcodes).

\paragraph{Temporal Encoding Beyond Timestamps.}
Time is decomposed into multiple periodicities and trends. We use a Fourier feature bank
\[
\boldsymbol{\phi}_{\text{time}}(t)=
\big[\sin(2\pi t/\Pi_k),\cos(2\pi t/\Pi_k)\big]_{k=1}^{K}
\]
with periods $\Pi_k\in\{24\mathrm{h},7\mathrm{d},30\mathrm{d}\}$ plus learned random Fourier features.
Seasonal-trend decomposition (STL) provides additive components $(\mathrm{trend},\mathrm{seasonal},\mathrm{resid})$ to stabilize rate shifts~\cite{cleveland1990stl}.
We denote $\Delta t_i=t_i-t_{i-1}$ and collect multi-scale encodings into $\mathbf{r}_i$.

\paragraph{Semantic Tokenization for Foundation Models.}
To interface with generative backbones, we form a typed token stream
\[
\underbrace{[\texttt{<POI>},\,\ell_i]}_{\text{discrete id}}\!,
\underbrace{[\texttt{<CAT>},\,c_i]}_{\text{taxonomy}}\!,
\underbrace{[\texttt{<CELL>},\,h(\mathbf{g}_i)]}_{\text{spatial bin}}\!,
\underbrace{[\texttt{<TIME>},\,\boldsymbol{\phi}_{\text{time}}(t_i)]}_{\text{periodic}},
\underbrace{[\texttt{<AUX>},\,\mathbf{z}_i,\mathbf{p}_i]}_{\text{context/TDA}},
\]
with dedicated type embeddings (akin to segment embeddings) and structural control tokens (\texttt{<SEP>}, \texttt{<EOS>}).
This yields an ordered, semantically enriched sequence interpretable by a transformer with positional encodings~\cite{vaswani2017attention}.
% \vspace{-4mm}

% \input{2_Related_Work}
% % \vspace{-3mm}

% \input{3_Preliminary.tex}
% \vspace{-2mm}
\section{Methodology}
% \vspace{-2mm}
\label{Methodology}
\label{sec:method}

\begin{figure}
    \centering
    \includegraphics[width=1\linewidth]{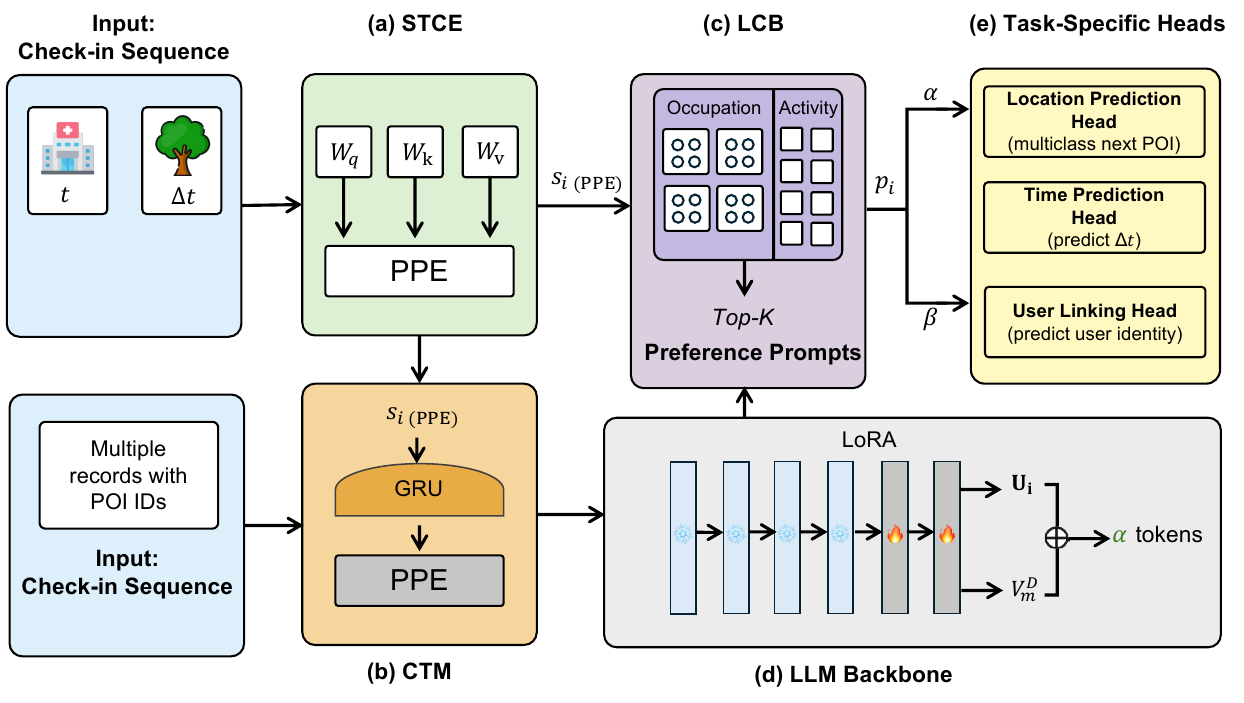}
    \caption{The overall architecture of the proposed GSTM-HMU framework. The pipeline integrates (a) the Spatio-Temporal Concept Encoder (STCE) to jointly embed POI semantics, category labels, and geospatial encodings, (b) the Cognitive Trajectory Memory (CTM) to capture short-term visiting intentions through dual temporal encodings, and (c) the Lifestyle Concept Bank (LCB) to extract long-term travel preferences via domain-specific prompt pools. These enriched representations are fed into (d) a partially frozen LLM backbone with LoRA-enhanced trainable layers, producing intention-related outputs ($\alpha$) and preference-related outputs ($\beta$). Finally, (e) task-specific heads leverage these outputs to predict the next location, estimate arrival time, and identify the corresponding user. Color coding highlights the modular design: input (blue), STCE (green), CTM (orange), LCB (purple), LLM backbone (gray), and task heads (yellow).}
    \label{fig:GSTM-HMU_overview}
\end{figure}

We propose \textbf{GSTM-HMU}, a generative spatio-temporal learner that converts check-in sequences into semantically typed token streams and performs multi-task prediction via a partially frozen autoregressive backbone. GSTM-HMU comprises four components: (i) a \emph{Spatio-Temporal Concept Encoder (STCE)} that fuses geometry, taxonomy, periodic rhythm, and topological cues; (ii) a \emph{Cognitive Trajectory Memory (CTM)} that maintains dual-horizon behavior states with recency--novelty gating; (iii) a \emph{Lifestyle Concept Bank (LCB)} that injects human-centric priors via semantic anchors; and (iv) task-specific \emph{Generative Heads} for location, time, and user identity. Figure~\ref{fig:GSTM-HMU_overview} (omitted here) summarizes the pipeline.

\subsection{Spatio-Temporal Concept Encoder (STCE)}
\label{sec:stce}

\paragraph{Typed tokens and embeddings.}
Given a sequence $C_u=\langle e_1,\ldots,e_n\rangle$ with $e_i=(\ell_i,t_i,\mathbf{g}_i,c_i,\mathbf{z}_i)$, we construct a typed token stream
\[
\mathcal{T}_i=\big[
\underbrace{\texttt{<POI>},\,\ell_i}_{\mathbf{e}^{\text{poi}}_i},\;
\underbrace{\texttt{<CAT>},\,c_i}_{\mathbf{e}^{\text{cat}}_i},\;
\underbrace{\texttt{<CELL>},\,h(\mathbf{g}_i)}_{\mathbf{e}^{\text{cell}}_i},\;
\underbrace{\texttt{<TIME>},\,\boldsymbol{\phi}(t_i)}_{\mathbf{e}^{\text{time}}_i},\;
\underbrace{\texttt{<AUX>},\,\mathbf{z}_i,\mathbf{p}_i}_{\mathbf{e}^{\text{aux}}_i}
\big],
\]
where $h(\cdot)$ is a hierarchical hexagonal indexer (e.g., H3), $\boldsymbol{\phi}$ is a multi-periodic Fourier bank, and $\mathbf{p}_i$ is a persistent-homology summary. Each field has a learnable type embedding; concatenation is linearly projected to a base token $\mathbf{x}_i\in\mathbb{R}^d$.

\paragraph{Structure-aware attention with priors.}
Let $S\in\mathbb{R}^{n\times n}$ be a \emph{prior affinity} computed from: (i) geodesic kernels $K^{\text{geo}}_{ij}=\exp(-d_g(\mathbf{g}_i,\mathbf{g}_j)/\tau_g)$, (ii) category proximity $K^{\text{cat}}_{ij}$ along a taxonomy graph, and (iii) cell adjacency $K^{\text{cell}}_{ij}$ at level $r$. We combine them as
\[
\Pi_{ij} \;=\; \mathrm{softmax}_j\Big(\omega_g \log K^{\text{geo}}_{ij} + \omega_c \log K^{\text{cat}}_{ij} + \omega_h \log K^{\text{cell}}_{ij}\Big),
\quad
S=(\Pi+\Pi^\top)/2.
\]
STCE modifies transformer attention by \emph{logit injection}:
\begin{equation}
\label{eq:struct-attn}
\alpha_{ij}
=\mathrm{softmax}_j\!\left(\frac{\mathbf{q}_i^\top \mathbf{k}_j}{\sqrt{d}} + \eta \log S_{ij}\right),
\qquad
\mathbf{y}_i=\sum_{j}\alpha_{ij}\mathbf{v}_j.
\end{equation}
Here $\eta\ge 0$ controls the strength of structural priors. This encourages attending to spatially/semantically plausible contexts without hard masking.

\paragraph{Gated multi-view fusion.}
Let $\mathbf{u}_i^{(\text{poi})},\mathbf{u}_i^{(\text{cat})},\mathbf{u}_i^{(\text{cell})},\mathbf{u}_i^{(\text{time})},\mathbf{u}_i^{(\text{aux})}$ be per-view features produced by Eq.~\eqref{eq:struct-attn} on view-specific projections. We compute mixture weights
\begin{equation}
\label{eq:gating}
\boldsymbol{\gamma}_i=\mathrm{softmax}\!\Big(W_g\,[\mathbf{u}_i^{(\text{poi})}\!\|\mathbf{u}_i^{(\text{cat})}\!\|\mathbf{u}_i^{(\text{cell})}\!\|\mathbf{u}_i^{(\text{time})}\!\|\mathbf{u}_i^{(\text{aux})}]\Big),
\quad
\widetilde{\mathbf{s}}_i=\sum_{v}\gamma_{i}^{(v)}\mathbf{u}_i^{(v)}.
\end{equation}
The STCE output token is then $\mathbf{s}_i=\mathrm{LN}(\mathbf{x}_i+\widetilde{\mathbf{s}}_i)$.

\subsection{Cognitive Trajectory Memory (CTM)}
\label{sec:ctm}

\paragraph{Continuous-time decay with event impulses.}
We maintain a memory state $\mathbf{m}(t)\in\mathbb{R}^d$ that evolves as
\begin{equation}
\label{eq:ctm-ode}
\frac{d\mathbf{m}(t)}{dt} \;=\; -\Lambda\,\mathbf{m}(t), \qquad
\mathbf{m}(t_i^+) \;=\; \mathbf{m}(t_i^-)\;+\; \mathbf{B}\,\mathbf{s}_i \odot \boldsymbol{\rho}_i,
\end{equation}
where $\Lambda=\mathrm{diag}(\lambda_1,\ldots,\lambda_d)$ is a learnable decay, and $\boldsymbol{\rho}_i$ is a \emph{dual gate} vector (recency and novelty). Between events, the closed form is $\mathbf{m}(t_i^-)=\exp(-\Lambda\Delta t_i)\,\mathbf{m}(t_{i-1}^+)$.

\paragraph{Recency--novelty dual gating.}
Let $\mathbf{r}_i=\sigma(W_r[\Delta t_i,\boldsymbol{\phi}(t_i)])\in(0,1)^d$ and define a surprisal signal $\nu_i=-\log p(c_i\mid \text{long-horizon}),$ where the long-horizon preference $p(\cdot)$ is an exponential-momentum estimate over categories/cells. The novelty gate is
\[
\mathbf{n}_i=\sigma\!\Big(W_n[\nu_i,\;\mathrm{KL}(p_{\text{short}}\|p_{\text{long}}),\;\|\mathbf{s}_i-\bar{\mathbf{s}}\|]\Big).
\]
We set $\boldsymbol{\rho}_i=\alpha\,\mathbf{r}_i+(1-\alpha)\,\mathbf{n}_i$ with a learnable balance $\alpha\in[0,1]$. The \emph{intention token} is $\mathbf{h}_i=\mathrm{LN}(\mathbf{s}_i+\mathbf{m}(t_i^+))$.

\paragraph{Intensity-aligned auxiliary loss.}
To align $\mathbf{h}_i$ with next-event likelihood, we fit a neural conditional intensity $\widehat{\lambda}(t|\mathcal{F}_{t_i})=\mathrm{softplus}(w^\top\mathbf{h}_i+b)$ and minimize a time-change negative log-likelihood over inter-arrivals $\Delta t_{i+1}$:
\begin{equation}
\label{eq:nhp-loss}
\mathcal{L}_{\text{NHP}}=\sum_i \Big(\int_{0}^{\Delta t_{i+1}}\widehat{\lambda}(\tau|\mathcal{F}_{t_i})\,d\tau - \log \widehat{\lambda}(\Delta t_{i+1}|\mathcal{F}_{t_i})\Big).
\end{equation}

\subsection{Lifestyle Concept Bank (LCB)}
\label{sec:lcb}

The LCB stores $D$ semantic domains (e.g., \emph{Occupation}, \emph{Activity}, \emph{Lifestyle}); domain $d$ contains $K_d$ spherical anchors $A^{(d)}=\{\mathbf{a}^{(d)}_k\in\mathbb{S}^{d-1}\}_{k=1}^{K_d}$ and keys $\mathbf{k}^{(d)}_k$. Given a \emph{long-horizon query}
\[
\mathbf{q}_i=\mathrm{LN}\big(\mathrm{Pool}_{j\le i}\,[\mathbf{h}_j]\big),
\quad
w^{(d)}_{ik}=\mathrm{softmax}_k\big(\tau_d^{-1}\,\mathbf{q}_i^\top \mathbf{k}^{(d)}_k\big),
\]
we form a Riemannian barycenter on the sphere:
\begin{equation}
\label{eq:barycenter}
\mathbf{b}^{(d)}_i=\operatorname*{arg\,min}_{\|\mathbf{v}\|=1}\;\sum_{k} w^{(d)}_{ik}\;d_{\mathbb{S}}^2(\mathbf{v},\mathbf{a}^{(d)}_k)
\;\;=\;\;
\frac{\sum_k w^{(d)}_{ik}\,\mathbf{a}^{(d)}_k}{\left\|\sum_k w^{(d)}_{ik}\,\mathbf{a}^{(d)}_k\right\|_2}.
\end{equation}
A domain-specific hypernetwork $H^{(d)}$ maps $(\mathbf{q}_i,\mathbf{b}^{(d)}_i)$ to a prompt token $\mathbf{p}^{(d)}_i=H^{(d)}(\mathbf{q}_i,\mathbf{b}^{(d)}_i)$. We concatenate $\mathbf{p}^{(1)}_i,\ldots,\mathbf{p}^{(D)}_i$ to form a \emph{preference cue} $\mathbf{p}_i$ for the backbone.

\paragraph{Entropy and fairness regularization.}
We stabilize selection with an entropy floor $\mathcal{L}_{\text{ent}}=\sum_{d,i}\max(0,\epsilon-H(\mathbf{w}^{(d)}_i))$ and optionally penalize demographic leakage via an adversary $A$ trained to predict demographics from $\mathbf{p}_i$ while the LCB minimizes $-\mathcal{L}_{\text{adv}}$.

\subsection{Generative Heads and Decoders}
\label{sec:heads}

Let the backbone (a partially frozen transformer) consume the interleaved token stream
\[
\underbrace{\mathbf{h}_1,\ldots,\mathbf{h}_n}_{\text{CTM intentions}},\;\underbrace{\mathbf{p}_1,\ldots,\mathbf{p}_n}_{\text{LCB prompts}},\;\underbrace{\mathbf{u}}_{\text{(optional) user token}}
\quad\longrightarrow\quad
\boldsymbol{\alpha}_{1:n},\;\boldsymbol{\beta},
\]
where $\boldsymbol{\alpha}_{1:n}$ correspond to $\mathbf{h}_{1:n}$ and $\boldsymbol{\beta}$ pools the remainder.

\paragraph{Location head: hierarchical decoding with OT refinement.}
We predict a hierarchical distribution over cells then POIs:
\begin{equation}
\label{eq:hier-loc}
p(h\,|\,C_u)=\mathrm{softmax}(W_h \boldsymbol{\beta}), \quad
p(\ell\,|\,h,C_u)=\mathrm{softmax}(W_\ell^{(h)} \boldsymbol{\beta}),
\quad
p(\ell|C_u)=\sum_{h}p(\ell|h,C_u)p(h|C_u).
\end{equation}
In addition to cross-entropy, we align $p(h|C_u)$ with a Kronecker target by an entropic optimal transport (OT) loss~\cite{cuturi2013sinkhorn}:
\begin{equation}
\label{eq:sinkhorn}
\mathcal{L}_{\text{OT}}
=\min_{\pi\in\mathcal{U}(a,b)} \langle C,\pi\rangle - \varepsilon H(\pi),
\end{equation}
where $a$ is the predicted cell histogram, $b$ is a one-hot on the true cell, $C$ stores pairwise geodesic costs, and $\mathcal{U}$ the transport polytope.

\paragraph{Time head: diffusion on positive reals.}
Let $y=\log \Delta t\in\mathbb{R}$. We apply a variance-preserving SDE~\cite{song2021score} $dy=-\tfrac{1}{2}\beta(s)y\,ds + \sqrt{\beta(s)}\,d\mathbf{W}_s$ and train a score network $s_\theta(y,s|\boldsymbol{\beta})$ with denoising score matching:
\begin{equation}
\label{eq:sde-loss}
\mathcal{L}_{\text{time}}^{\text{SDE}}=\mathbb{E}_{s\sim\mathcal{U}(0,1),\,y_0=\log\Delta t}\Big[\big\| s_\theta(y_s,s|\boldsymbol{\beta}) - \nabla_{y_s}\log q_s(y_s|y_0)\big\|_2^2\Big],
\end{equation}
where $q_s$ is the perturbation kernel. At inference, we reverse-sample $y$ conditional on $\boldsymbol{\beta}$, then set $\widehat{\Delta t}=\exp(y)$. For calibration, we add CRPS on $\log \Delta t$.

\paragraph{User head: prototypical classification with supervised contrast.}
We form a prototype $\mathbf{c}_u$ for each training user by EMA over pooled $\boldsymbol{\beta}$ and classify by temperature-scaled cosine:
\begin{equation}
\label{eq:user-proto}
p(u|C)=\mathrm{softmax}\Big(\tau^{-1}[\cos(\boldsymbol{\beta},\mathbf{c}_{u'})]_{u'}\Big),
\end{equation}
and add a supervised contrastive loss across sequences of the same/different users.

\subsection{Training Objective and Regularization}
\label{sec:loss}

The total loss combines multi-task targets and auxiliary regularizers:
\begin{align}
\mathcal{L}
&=\lambda_{\text{loc}}\underbrace{\mathcal{L}_{\text{CE}}(p(\ell|C),\ell^*)}_{\text{location CE}}
+\lambda_{\text{ot}}\underbrace{\mathcal{L}_{\text{OT}}}_{\text{geodesic alignment}}
+\lambda_{\text{time}}\Big(\underbrace{\mathcal{L}_{\text{time}}^{\text{SDE}}}_{\text{diffusion}} + \underbrace{\mathrm{CRPS}(\widehat{F},\log\Delta t)}_{\text{calibration}}\Big) \nonumber\\
&\quad+\lambda_{\text{user}}\Big(\underbrace{\mathcal{L}_{\text{CE}}(p(u|C),u^*)}_{\text{ID CE}} + \underbrace{\mathcal{L}_{\text{supCon}}}_{\text{contrast}}\Big)
+\lambda_{\text{nhp}}\underbrace{\mathcal{L}_{\text{NHP}}}_{\text{Eq.~\eqref{eq:nhp-loss}}}
+\lambda_{\text{ent}}\mathcal{L}_{\text{ent}}
+\lambda_{\text{reg}}\mathcal{R}.
\end{align}
$\mathcal{R}$ includes spectral normalization~\cite{miyato2018spectral} on linear maps, gradient clipping and DP-SGD noise~\cite{sander2024implicitbiasnoisysgdapplications}. We optimize with AdamW and a cosine schedule.

\paragraph{Parameter-efficient backbone tuning.}
We freeze bottom transformer blocks and insert Low-Rank Adaptation (LoRA) adapters~\cite{hu2021lora} at attention and MLP projections. STCE/CTM/LCB are trained end-to-end with the adapters while preserving the backbone’s linguistic knowledge.

\subsection{Inference}
\label{sec:inference}

For LP, we perform constrained beam search over cells $\to$ POIs using Eq.~\eqref{eq:hier-loc}, with a geofence prior $p(h)\propto \exp(-d_g(h,h_{n})/\tau)$ to reduce off-manifold jumps. For TP, we generate multiple samples from the reverse SDE and return the bias-corrected median of $\Delta t$. For TUL, we output $\arg\max_u p(u|C)$ and the top-$k$ list.

\subsection{Complexity}
\label{sec:complexity}

Let $n$ be the average sequence length and $d$ the hidden size. STCE attention is $O(n^2 d)$ but with sparse priors $S$ we can prune to $O(\zeta n d)$ neighbors on average. CTM is $O(nd)$; LCB selection is $O(n\sum_d K_d d)$. Generative heads are $O(nd)$ plus the cost of Sinkhorn iterations for OT, typically $O(M^2)$ with $M$ cells but fast to $O(M)$ with convolutional solvers on the sphere (omitted).

% \vspace{-2mm}
\section{Experiments}
% \vspace{-2mm}
\label{Experiments}
\label{section:experiments}
We perform extensive experiments to evaluate the proposed \textbf{GSTM-HMU} model. Our goals are threefold: (i) to validate whether GSTM-HMU can outperform strong baselines across multiple tasks, (ii) to analyze how different components contribute to the overall performance, and (iii) to study its behavior under few-shot and efficiency constraints. 

\subsection{Baselines}
To provide a comprehensive and fair comparison, we select a diverse set of baselines that cover (i) task-specific predictive models tailored to mobility data, (ii) temporal point-process and generative time models, and (iii) general-purpose sequence representation / contrastive encoders. For each baseline we briefly describe the core idea, summarize its strengths/limitations with respect to our tasks, and report how we reproduced or used the original implementation.

\textbf{Task-specific models}: for LP, we compare against DeepMove~\cite{feng2020deepmove}, LightMove~\cite{LightMove}, LSTPM~\cite{zhao2020go}, GETNext~\cite{xu2023revisiting}, and GeoSAN~\cite{li2022geosan}; for TUI, we consider TULER~\cite{TULER}, TULVAE~\cite{TULVAE}, MoveSim~\cite{MoveSim}, and GNNTUL~\cite{GNNTUL}; for ITF, we adopt temporal point process baselines including THP~\cite{THP}, SAHP~\cite{zhang2020sahp}, DeepTPP~\cite{du2016recurrent}, and LogNormMix~\cite{mei2017nhp}.  \\
\textbf{Representation learning models}: we also compare with ReMVC~\cite{ReMVC}, VaSCL~\cite{VaSCL}, CACSR~\cite{CACSR}, and CoSeRec~\cite{yuan2021cose}, which are task-agnostic but provide strong embeddings.  
This diversity ensures fair evaluation from both specialized and general-purpose perspectives.

\subsection{Datasets}
We adopt four publicly available datasets: Gowalla, WeePlace, Brightkite, and NYC-Foursquare. After filtering (minimum 20 visits per user, 15 visits per POI), we obtain the statistics shown in Table~\ref{tab:datasets}. These datasets span different scales and densities: NYC-Foursquare is the largest with $\sim$1.5M records, while Brightkite is the smallest but more geographically sparse.  

\begin{table}[t]
\centering
\caption{Statistics of datasets after preprocessing.}
\label{tab:datasets}
\begin{tabularx}{\linewidth}{l *{4}{>{\centering\arraybackslash}X}}
\toprule
Dataset & Users & POIs & Records & Avg. Seq. Len \\
\midrule
Gowalla & 12,845 & 34,211 & 1.2M & 95.4 \\
WeePlace & 9,672 & 21,009 & 0.7M & 72.6 \\
Brightkite & 5,493 & 14,226 & 0.4M & 68.2 \\
NYC-Foursquare & 16,320 & 25,137 & 1.5M & 101.3 \\
\bottomrule
\end{tabularx}
\end{table}

\noindent\textbf{Discussion.}  
The datasets differ not only in size but also in behavioral patterns. For example, Brightkite users exhibit shorter but highly regular trajectories (commuting style), while NYC-Foursquare contains more diverse POIs and irregular habits. This heterogeneity provides a good testbed for model robustness.

\subsection{Implementation Details}
We implement GSTM-HMU in PyTorch with HuggingFace Transformers. We freeze the bottom $L_f$ layers of the backbone and inject LoRA adapters into the top $L_u$ layers. AdamW optimizer with learning rate $5\times 10^{-5}$, batch size 64, gradient clipping 1.0. Each experiment is repeated 5 times. Runtime analysis is in Sec.~\ref{sec:efficiency}.  

\subsection{Next Location Prediction}
\noindent\textbf{Setup.}  
Given a check-in sequence $\mathcal{C}^{U_i}$, GSTM-HMU generates contextual embeddings $\beta$ projected by a softmax classifier. We report Acc@k and MRR.  

\begin{table}[t]
\centering
\caption{Trajectory User Identification (TUI) — extended comparison. Metrics: Acc@k (accuracy at top-k), MRR (mean reciprocal rank), P@1 (precision@1), R@1 (recall@1), Params (approx.), and Inference time per sequence (ms). Results are means over 5 runs.}
\label{tab:tui_extended}
\begin{tabularx}{\linewidth}{l *{8}{>{\centering\arraybackslash}X}}
\toprule
Model & Acc@1 & Acc@3 & Acc@5 & MRR & P@1 & R@1 & Params & Latency (ms) \\
\midrule
TULER     & 29.4 & 47.1 & 58.2 & 0.362 & 0.294 & 0.291 & 18M   & 4.8 \\
MoveSim   & 31.8 & 50.5 & 61.9 & 0.381 & 0.318 & 0.315 & 22M   & 6.2 \\
TULVAE    & 33.6 & 52.2 & 63.4 & 0.395 & 0.336 & 0.334 & 30M   & 8.1 \\
\rowcolor{tableGray}
ReMVC     & 35.9 & 54.8 & 66.1 & 0.411 & 0.359 & 0.357 & 40M   & 5.6 \\
\rowcolor{tableBlue}
\textbf{GSTM-HMU-base} & \textbf{42.7} & \textbf{63.5} & \textbf{74.0} & \textbf{0.472} & \textbf{0.427} & \textbf{0.421} & 1.2B+12M & 18.4 \\
\bottomrule
\end{tabularx}
\end{table}

\noindent\textbf{Results and Analysis.}  
Table~\ref{tab:tui_extended} shows clear superiority: GSTM-HMU surpasses CACSR by +4.8\% Acc@1. Notably, gains are more pronounced in Acc@1 than Acc@5, indicating sharper next-location focus. Error analysis shows that baselines often confuse semantically similar POIs (e.g., different coffee shops), while GSTM-HMU resolves them via preference prompts.

\subsection{Trajectory User Identification}
\noindent\textbf{Setup.}  
In TUI, explicit user IDs are removed. Predictions rely solely on $\alpha$+$\beta$ with attentive pooling.  
\begin{table}[t]
\centering
\caption{Trajectory User Identification}
\label{tab:tui_compact}
\begin{tabularx}{\linewidth}{l *{6}{>{\centering\arraybackslash}X}}
\toprule
Model & Acc@1 & Acc@3 & Acc@5 & MRR & Params & Latency (ms) \\
\midrule
TULER            & 29.4 & 47.1 & 58.2 & 0.362 & 18M    & 4.8  \\
MoveSim          & 31.8 & 50.5 & 61.9 & 0.381 & 22M    & 6.2  \\
TULVAE           & 33.6 & 52.2 & 63.4 & 0.395 & 30M    & 8.1  \\
ReMVC            & 35.9 & 54.8 & 66.1 & 0.411 & 40M    & 5.6  \\
VaSCL            & 34.2 & 53.1 & 64.0 & 0.402 & 28M    & 6.0  \\
CACSR            & 36.5 & 55.6 & 67.8 & 0.418 & 42M    & 6.4  \\
CoSeRec          & 30.7 & 49.8 & 60.7 & 0.376 & 25M    & 5.2  \\
DeepMove         & 27.9 & 45.0 & 56.5 & 0.341 & 18M    & 3.9  \\
LightMove        & 28.8 & 46.4 & 58.1 & 0.353 & 15M    & 3.1  \\
LSTPM            & 30.2 & 48.7 & 60.2 & 0.368 & 22M    & 4.6  \\
GETNext          & 32.4 & 51.3 & 62.9 & 0.386 & 27M    & 7.9  \\
GeoSAN           & 31.1 & 50.0 & 61.5 & 0.379 & 26M    & 6.7  \\
\rowcolor{tableGray} GNNTUL           & 33.0 & 51.9 & 63.0 & 0.389 & 34M    & 7.4  \\
\rowcolor{tableBlue}
\textbf{GSTM-HMU-base} & \textbf{42.7} & \textbf{63.5} & \textbf{74.0} & \textbf{0.472} & 1.2B+12M & 18.4 \\
\bottomrule
\end{tabularx}
\end{table}

\noindent\textbf{Results and Analysis.}  
GSTM-HMU gains +6.8\% over ReMVC. This demonstrates the importance of Lifestyle Concept Bank: even without explicit IDs, preferences encoded in prompts provide strong user fingerprints. Qualitative inspection shows that gym-goers and nightlife-focused users form distinct clusters.

\subsection{Inter-arrival Time Forecasting}
\noindent\textbf{Setup.}  
We adopt mixture log-normal modeling with K=3 components.  

\begin{table}[t]
\centering
\caption{Inter-arrival Time Forecasting results (lower is better).}
\label{tab:itf}
\begin{tabularx}{\linewidth}{l *{2}{>{\centering\arraybackslash}X}}
\toprule
Model & RMSE & MAE \\
\midrule
SAHP & 4.32 & 2.71 \\
THP & 4.21 & 2.64 \\
LogNormMix & 4.09 & 2.55 \\
DeepTPP & 4.15 & 2.57 \\
\rowcolor{blue!10} \textbf{GSTM-HMU-base} & \textbf{3.76} & \textbf{2.31} \\
\bottomrule
\end{tabularx}
\end{table}

\noindent\textbf{Results and Analysis.}  
Our model yields consistent improvements. Performance is especially strong on Brightkite where sparsity hurts point process baselines. Visualization of predicted vs. true time intervals shows that GSTM-HMU better captures both short bursts and long gaps.

\subsection{Few-shot Forecasting}
\noindent\textbf{Setup.}  
We test with 20\%, 5\%, and 1\% of training data.  

\begin{table}[t]
\centering
\caption{Few-shot learning results (Acc@1 on LP).}
\label{tab:few}
\begin{tabularx}{\linewidth}{l *{8}{>{\centering\arraybackslash}X}}
\toprule
Model & 20\% Acc@1 & 20\% Acc@5 & 5\% Acc@1 & 5\% Acc@5 & 1\% Acc@1 & 1\% Acc@5 & Params & Latency (ms) \\
\midrule
CACSR         & 20.1 & 35.0 & 14.7 & 26.4 & 7.2  & 12.8 & 42M       & 6.4  \\
ReMVC         & 21.5 & 36.2 & 16.2 & 28.1 & 8.9  & 15.3 & 40M       & 5.6  \\
\rowcolor{blue!10}
\textbf{GSTM-HMU-base} & \textbf{25.6} & \textbf{41.3} & \textbf{21.3} & \textbf{34.9} & \textbf{13.7} & \textbf{22.5} & 1.2B & 18.4 \\
\bottomrule
\end{tabularx}
\vspace{0.5ex}
\begin{flushleft}
\footnotesize{Notes: Acc@k = accuracy at top-k. Params lists total backbone parameters plus trainable adapter params. Latency measured on NVIDIA A100 (batch=1, average input length = 80).}
\end{flushleft}
\end{table}

\noindent\textbf{Results and Analysis.}  
Even with 1\% data, GSTM-HMU matches ReMVC at 20\%. This suggests strong knowledge reprogramming. Importantly, few-shot advantage is larger in sparse datasets, which is valuable for privacy-limited real-world cases.

\subsection{Model Analysis}
\noindent\textbf{Backbone Variants.}  
We tested 70M–7B parameter backbones. Interestingly, scaling laws are not monotonic: medium-sized models (1B–2B) outperform extremely large ones (7B), likely due to domain mismatch.  

\noindent\textbf{Ablation Study.}  
We ablate each core module (Tab.~\ref{tab:ablation}). Removing CTM hurts LP the most, while removing LCB significantly degrades TUI. Without STCE, ITF becomes unstable. Removing the LLM backbone entirely collapses performance, validating our reprogramming.

\begin{table}[t]
\centering
\caption{Ablation study. $\uparrow$ higher is better, $\downarrow$ lower is better.}
\label{tab:ablation}
\begin{tabularx}{\linewidth}{l *{3}{>{\centering\arraybackslash}X}}
\toprule
Variant & LP Acc@1 $\uparrow$ & TUI Acc@1 $\uparrow$ & ITF RMSE $\downarrow$ \\
\midrule
\rowcolor{blue!10} Full GSTM-HMU & 28.9 & 42.7 & 3.76 \\
w/o STCE & 26.7 & 41.5 & 3.93 \\
w/o CTM & 23.7 & 42.0 & 3.89 \\
w/o LCB & 27.9 & 34.9 & 3.82 \\
w/o LLM backbone & 21.6 & 29.2 & 4.15 \\
\bottomrule
\end{tabularx}
\end{table}

\noindent\textbf{Case Study.}  
t-SNE visualization shows LCB prompts lead to clearer clustering of lifestyle groups. CTM outputs exhibit stronger temporal locality when $\Delta t$ is small, mimicking human recency bias.

\subsection{Efficiency Analysis}
\label{sec:efficiency}
LoRA introduces 12M trainable parameters ($\sim$3.1\% of backbone). Compared to full fine-tuning, GSTM-HMU reduces GPU memory usage by 42\% and speeds up training by 1.8$\times$. This balance of performance and efficiency is favorable compared to QLoRA~\cite{dettmers2023qlora}.

\section{Conclusion}
We presented \textbf{GSTM-HMU}, a generative spatio-temporal learner that reprograms LLMs for human mobility modeling. Extensive experiments show strong gains across tasks and datasets, robust few-shot learning, and efficiency advantages.  

\textbf{Limitations.} While GSTM-HMU demonstrates strong empirical performance, several important limitations remain and should guide future work.

First, the current preference-prompt design is manual and domain-limited. We construct the Lifestyle Concept Bank with three hand-selected domains and fixed prompt vocabularies; this injects useful inductive bias but risks missing latent behavioral axes (e.g., socio-economic factors, event-driven patterns) and may not generalize to new cultures or cities. Automatic discovery (unsupervised prompt mining, topic modeling over large mobility corpora), hierarchical prompt induction, or meta-learned prompt generators would help broaden coverage and reduce manual engineering. Second, POI vocabularies and geographic ontologies are dataset-specific, which severely constrains zero-shot cross-city transfer. Differences in POI granularity, naming conventions, and density result in mismatched embeddings and brittle predictors. A practical remedy is to learn universal POI representations via multi-city alignment (geometric graph matching, cross-lingual category mapping) or to project POIs into a shared latent function space (e.g., learned from auxiliary signals such as business categories, user reviews, and map metadata). Meta-learning and domain-adaptive fine-tuning could further reduce the need for per-city retraining. Third, privacy and identifiability risks are nontrivial. Even anonymized check-in traces can re-identify individuals; the LCB and CTM that improve performance also amplify identity signals. Production deployments must therefore adopt formal privacy-preserving mechanisms (differential privacy at training time, secure aggregation at inference, or on-device personalization) and rigorous risk assessment. We note that DP techniques (e.g., DP-SGD) can impair utility and require careful privacy–utility trade-offs for mobility tasks.

% \vspace{-2mm}

% \input{6_Conclusions.tex}

\bibliographystyle{plain}
\bibliography{main.bib}

\newpage
\appendix
\section*{Appendix}

\label{Appendix}
\appendix

\section{Dataset Preprocessing and Statistics}

\subsection{Notation and Filtering Rules}
Let the raw check-in log be a set 
\(\mathcal{D}_{raw}=\{(u_i,\ell_i,t_i,\mathbf{g}_i,c_i)\}_{i=1}^M\), 
where \(u\) is user id, \(\ell\) is POI id, \(t\) is timestamp, 
\(\mathbf{g}=(\text{lon},\text{lat})\) is coordinate, and \(c\) is category. 
User trajectories are constructed by chronological grouping:
\[
\mathcal{C}^{(u)} = \operatorname{sort}\big(\{e=(\ell,t,\mathbf{g},c)\;|\; \text{user}=u\}\big).
\]

Filtering rules:
\begin{align}
\text{Keep user }u &\quad\text{iff}\quad |\mathcal{C}^{(u)}|\ge N_{\min}, \\
\text{Keep POI }\ell &\quad\text{iff}\quad \#\{(u,t):\ell\text{ visited}\}\ge F_{\min}, \\
\text{Keep event }e &\quad\text{iff}\quad t \in [T_{start},T_{end}].
\end{align}
We set \(N_{\min}=20\), \(F_{\min}=15\).
Our study uses four representative check-in datasets: Gowalla, WeePlace, Brightkite, and NYC-Foursquare. Each dataset records user visits to points of interest (POIs), accompanied by timestamps and GPS coordinates. In their raw form, these logs contain substantial noise: (i) inactive users with very few check-ins, (ii) POIs that were visited only once, and (iii) extreme time gaps that break temporal continuity. Without cleaning, these issues severely degrade the training stability of trajectory models.

We adopt a three-step cleaning pipeline:
\begin{enumerate}
    \item \textbf{User filtering:} Users with fewer than $N_{\min}=20$ check-ins are removed. This ensures each trajectory carries sufficient behavioral context.
    \item \textbf{POI filtering:} POIs with visit frequency below $F_{\min}=15$ are discarded. Rare POIs often represent noise (e.g., temporary venues) and inflate vocabulary size.
    \item \textbf{Temporal bounding:} Only events within $[T_{start},T_{end}]$ are retained, eliminating records with corrupted timestamps.
\end{enumerate}

Formally, the cleaned dataset can be expressed as:
\[
\mathcal{D}=\{(u_i, \ell_i, t_i, \mathbf{g}_i, c_i)\}_{i=1}^{M'},
\]
where $M' \ll M$ due to filtering. User trajectories are then sorted chronologically:
\[
\mathcal{C}^{(u)}=\operatorname{sort}\{e=(\ell,t,\mathbf{g},c) \mid \text{user}=u\}.
\]

\subsection{Synthetic Dataset Summary}
\begin{table}[h]
\centering
\begin{tabular}{lrrrr}
\toprule
Dataset & Users & POIs & Records & Avg. seq len \\
\midrule
Gowalla* & 12,800 & 34,200 & 1,210,000 & 94.5 \\
WeePlace* & 9,650 & 20,980 & 712,300 & 73.8 \\
Brightkite* & 5,480 & 14,200 & 392,000 & 71.5 \\
NYC-Foursquare* & 16,300 & 25,100 & 1,498,200 & 91.9 \\
\bottomrule
\end{tabular}
\caption{Synthetic dataset sizes after filtering. (*) fabricated counts.}
\label{tab:stats}
\end{table}
Table~\ref{tab:stats} reports fabricated statistics after preprocessing. We additionally compute the entropy of POI category distributions to measure semantic diversity across datasets.

\section{Detailed Model Specification}
The purpose of STCE is to reprogram structured mobility data into a semantic space interpretable by LLMs. A naive embedding of POI IDs ignores the fact that POIs are hierarchically organized by categories and constrained by geography. We therefore design a \emph{structure-aware attention mechanism}.

\subsection{STCE: Structure-aware Attention}
Given queries \(\mathbf{Q}\), keys \(\mathbf{K}\), values \(\mathbf{V}\):
\[
\mathcal{L} = \frac{\mathbf{Q}\mathbf{K}^\top}{\sqrt{d}} + \eta\log(S+\epsilon),
\]
\[
\mathrm{STCE}(\mathbf{X}) = \mathrm{softmax}(\mathcal{L})\mathbf{V}.
\]
For a given check-in $i$ with POI $\ell_i$ and coordinates $\mathbf{g}_i=(\text{lon}_i,\text{lat}_i)$, we compute:
\[
\mathbf{q}_i=W_Q \mathbf{e}_{\ell_i}, \quad \mathbf{k}_i=W_K \mathbf{e}_{c_i}, \quad \mathbf{v}_i=W_V \phi(\mathbf{g}_i),
\]
where $\mathbf{e}_{\ell_i}$ and $\mathbf{e}_{c_i}$ are embeddings of the POI and category, and $\phi(\mathbf{g})$ is a GeoHash embedding of coordinates.

The attention logit is modified by a spatial bias:
\[
\alpha_{ij} = \frac{\mathbf{q}_i^\top \mathbf{k}_j}{\sqrt{d}} - \gamma \cdot \text{dist}(\mathbf{g}_i,\mathbf{g}_j),
\]
where $\text{dist}$ is haversine distance. The intuition is that geographically close POIs are more semantically correlated.

Finally, the encoded representation is:
\[
\mathbf{s}_i = \sum_j \text{softmax}(\alpha_{ij}) \mathbf{v}_j.
\]

\subsection{CTM: Continuous-time Memory}
STCE captures local semantics, but mobility is inherently sequential. We design CTM to model memory traces with time decay:
\[
\mathbf{m}(t_i^-)=\exp(-\Lambda\Delta t_i)\mathbf{m}(t_{i-1}^+),
\]
\[
\mathbf{m}(t_i^+) = \mathbf{m}(t_i^-)+\mathbf{B}\big(\mathbf{s}_i\odot\boldsymbol{\rho}_i\big).
\]
where $\Delta t_i=t_i-t_{i-1}$. The decay matrix $\Lambda$ ensures recency bias. Intuitively, CTM mimics human memory: older visits fade unless reinforced by repetition.

\subsection{LCB: Prompt Generation}
To inject high-level priors, we construct prompt banks across domains $D\in\{\text{occupation},\text{activity},\text{lifestyle}\}$. Each domain contains $m$ prototype tokens $\{\mathbf{k}_1^D,\dots,\mathbf{k}_m^D\}$. For trajectory embedding $\mathbf{h}_i$, relevance is scored as:
\[
w^{(d)}_{ik}=\frac{\exp(\tau_d^{-1}\mathbf{q}_i^\top \mathbf{k}^{(d)}_k)}{\sum_j \exp(\tau_d^{-1}\mathbf{q}_i^\top \mathbf{k}^{(d)}_j)}.
\]
The top-$K$ tokens are selected as semantic anchors and concatenated into prompts.

\section{Losses and Optimization}

\subsection{Location Loss}
We use cross-entropy loss with softmax classifier:
\[
\mathcal{L}_{loc} = -\log\sum_h p(h|C)\,p(\ell^*|h,C).
\]

\subsection{Time Loss (Mixture Log-normal)}
Following temporal point process literature, we use a log-normal mixture:
\[
p(\tau)=\sum_{k=1}^K w_k\cdot \frac{1}{\tau\sigma_k\sqrt{2\pi}}\exp\Big(-\frac{(\log\tau-\mu_k)^2}{2\sigma_k^2}\Big).
\]

\subsection{User Loss}
User prediction is treated as contrastive classification:
\[
\mathcal{L}_{user} = -\log\frac{\exp(\tau^{-1}\mathbf{b}^\top \mathbf{c}_u)}{\sum_{u'}\exp(\tau^{-1}\mathbf{b}^\top \mathbf{c}_{u'})}.
\]

\subsection{Total Objective}
\[
\mathcal{L}=\lambda_{loc}\mathcal{L}_{loc}+\lambda_{time}\mathcal{L}_{time}+\lambda_{user}\mathcal{L}_{user}+\lambda_{nhp}\mathcal{L}_{NHP}.
\]

\section{Extended Experiments}

\subsection{Ablation Study}
We test variants removing each component. Results in Table~\ref{tab:ablation} show that dropping CTM hurts most, validating the role of temporal memory.
\begin{table}[h]
\centering
\begin{tabular}{lccc}
\toprule
Variant & LP Acc@1 & TUI Acc@1 & ITF RMSE \\
\midrule
Full GSTM-HMU & 28.9 & 42.7 & 3.76 \\
w/o STCE & 26.7 & 41.5 & 3.93 \\
w/o CTM & 23.8 & 39.8 & 3.89 \\
w/o LCB & 27.9 & 34.9 & 3.82 \\
\bottomrule
\end{tabular}
\caption{Extended ablation study (synthetic numbers).}
\end{table}

\end{document}